\documentclass{article}

\usepackage{arxiv}

\usepackage[utf8]{inputenc} % allow utf-8 input
\usepackage[T1]{fontenc}    % use 8-bit T1 fonts
\usepackage{hyperref}       % hyperlinks
\usepackage{url}            % simple URL typesetting
\usepackage{booktabs}       % professional-quality tables
\usepackage{amsfonts}       % blackboard math symbols
\usepackage{nicefrac}       % compact symbols for 1/2, etc.
\usepackage{microtype}      % microtypography
\usepackage{lipsum}
\usepackage{graphicx}
\graphicspath{ {./images/} }

\usepackage{cite}
\usepackage{times}
\usepackage{latexsym}
\usepackage[most]{tcolorbox}
\usepackage[T1]{fontenc}
\usepackage{fancyvrb}
\usepackage{algorithm} 
\usepackage{algpseudocode} 
\usepackage{soul}
\usepackage{arydshln}
\usepackage{xcolor}
\usepackage{booktabs}
\usepackage{multirow}
\usepackage{subcaption}
\usepackage[symbol]{footmisc}

\title{DoPAMine: Domain-specific Pre-training Adaptation from seed-guided data Mining}

\author{
Vinayak Arannil$^*$, Neha Narwal$^*$, Sourav Sanjukta Bhabesh$^*$,\\
\textbf{Sai Nikhil Thirandas, Darren Yow-Bang Wang, Graham Horwood, Alex Anto Chirayath, Gouri Pandeshwar} \\
\centerline{AWS AI}\\
\centerline{\texttt{\{varannil, sbhabesh, nnarwal, saintamz, ybwang, ghorwood, alexchir, gpandesh\}@amazon.com}}
}

\begin{document}
\maketitle
\begin{abstract}
Large Language Models (LLMs) have shown remarkable ability to generalize effectively across numerous industry domains while executing a range of tasks. Many of these competencies are obtained from the data utilized during the pre-training phase of the Language Models (LMs). However, these models exhibit limitations when tasked with performing in specialized or low-resource industry domains. More recent approaches use LLMs for generating domain-specific synthetic data but most often they lack in truthfulness and complexity. Alternatively, in cases where domain data is available like healthcare and finance most of the LMs are proprietary necessitating the need for a scalable method to curate real world industry specific pre-training data. In this work, we propose an automated and scalable framework - \textbf{DoPAMine}:\textbf{Do}main-specific \textbf{P}re-training \textbf{A}daptation from seed-guided data \textbf{Min}ing, to mine domain specific training data from a large data corpus for domain adaptation of a LM. The framework leverages the parametric knowledge of a LLM to generate diverse and representative seed data tailored to a specific domain which is then used to mine real world data from a large data corpus like Common Crawl. We evaluated our framework's performance in the continual pre-training (CPT) setting by training two domain specific 7B parameter LMs in healthcare and finance with data mined via DoPAMine. Our experiments show that DoPAMine boosts the performance of pre-trained LLMs on average by 4.9\% and 5.1\% in zero-shot and 5-shot settings respectively on healthcare tasks from MMLU, MedQA, MedMCQA and PubMedQA datasets, and 2.9\% and 6.7\% for zero-shot and 5-shot settings respectively on finance tasks from FiQA-SA, FPB and Headlines datasets when compared to the baseline.
\end{abstract}

% keywords can be removed

\footnotetext[1]{These authors contributed equally to this work.}
\footnotetext[7]{Preprint. Under review.}

\section{Introduction}

\begin{figure*}%[!thb]
%   \centering
  \includegraphics[width=\textwidth]{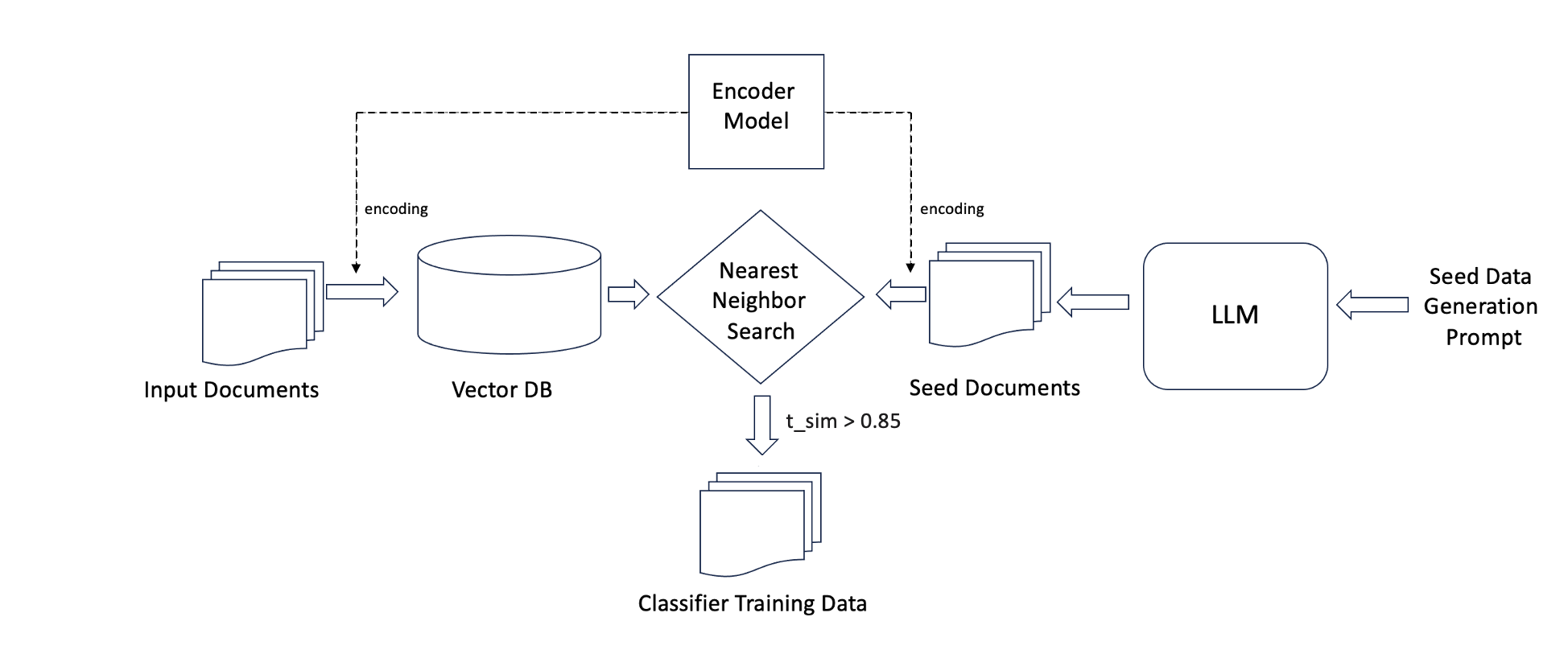}
  \caption{DoPAMine components for automated and scalable mining of domain specific data}
  \label{fig:myplot}
\end{figure*}

The rapid advancement of large language models (LLMs) has ushered in a new era of natural language processing, enabling remarkable capabilities in tasks such as text generation, summarization, and question answering \cite{NEURIPS2020_6b493230, brown2020language,thoppilan2022lamda,openai2023gpt, chowdhery2023palm, rae2021scaling, hoffmann2022training, jin2024comprehensive, kim2023sure}. Significant advancements have been made in training large language models which cater to multiple domains by training over a large corpus of data across domains \cite{touvron2023llama, hoffmann2022training, ouyang2022training, pmlr-v162-borgeaud22a}. Alternatively, smaller models targeting a specific task have been trained using synthetic data \cite{gunasekar2023textbooks,eldan2023tinystories}, specializing smaller models for a given task using multi-step reasoning \cite{pmlr-v162-borgeaud22a, pmlr-v202-fu23d, arannil-etal-2023-adeqa} and using instruction fine tuning for improving the LMs \cite{ding2023enhancing}.\par

The development of domain-specific language models is crucial for a wide range of applications, including but not limited to medical information extraction, financial report summarization, and legal document analysis. These domains often involve specialized vocabularies, writing styles, and linguistic patterns that require models to be trained on relevant data to achieve optimal performance. Relying solely on general-purpose language models trained on broad web data may result in suboptimal performance and a lack of domain-specific semantic understanding \cite{pal2024domain}. However, the success of these models hinges on the availability of high-quality, domain-specific training data, which remains a significant challenge \cite{zhang2023reformulating}.  Also, while there is an abundance of raw, uncategorized data available from web crawls and other sources, curated datasets tailored for pre-training LLMs in specific industry domains are scarce.\par

To address this challenge, we propose an automated and scalable framework - DoPAMine for mining domain-specific training data from large data corpora, such as Common Crawl \cite{dodge2021documenting}. Our approach leverages the power of large language models (LLMs), combined with embedding techniques and nearest neighbor search, to identify and extract relevant data for a given target domain. The core of our framework involves prompting a LLM, such as Claude 3 Sonnet \cite{claudev3}, using a chain-of-thought \cite{wei2023chainofthought} based prompt template tailored for generating domain-specific synthetic seed data. We then mine the large data corpus (in this paper we will use web crawl data) for semantically similar documents. The components of the DoPAMine framework are shown in figure \ref{fig:myplot}. 
We make the following contributions in this paper:
\begin{itemize}
  \item We introduce DoPAMine framework that is truly scalable and allows full control over the industry domains being mined. Unlike alternative approaches, such as clustering the entire data corpus, which face challenges like scalability, manual cluster inspection, and potential failure to produce clean clusters for low-resource domains, our framework offers precise control through synthetically generated seed data, ensuring the retrieval of semantically similar documents from the data corpus for the target domain.
  \item We propose a versatile prompt template that harnesses the parametric knowledge of a large language model (LLM) to generate diverse and representative seed data tailored to a specific industry domain. By carefully crafting the prompt template and varying factors such as document types, personas, author demeanors, intended audiences, and generation lengths, we create a rich tapestry of seed data that serves as a comprehensive exemplar for the target domain.
  \item Moving away from relying on fully synthetic data for domain adaptation, which often grapples with factuality and hallucination issues, our framework curates real-world data for the domain, fostering a more authentic and reliable representation of the target industry.
  \item Through an extensive ablation study, we show measurable improvements in the performance of downstream domain-specific tasks when the data curated through our DoPAMine framework is incorporated into the continued pre-training regime of LLMs.
\end{itemize}

In the following sections, we provide a detailed description of our methodology, experimental setup, and evaluation results, demonstrating the effectiveness of our automated data mining framework DoPAMine in curating domain-specific training data for LLMs.

\section{Methodology}

\begin{algorithm*}[!ht]
\caption{Seed guided automated data curation}
\label{algo1}
\begin{algorithmic}[1]
\Require Input Documents $D$, Encoder Model $E$, Seed Data Generation Prompt $P$, LLM $L$, Count of Seed Documets $C$
\Ensure Classifier Training Data $T$

\State $V \gets \{\}$ \Comment{Initialize Vector Database}
\For{$d \in D$}
    \State $v \gets E.encode(d)$ \Comment{Encode document using Encoder Model}
    \State $V \gets V \cup \{v\}$ \Comment{Add vector to Vector Database}
\EndFor

\State $S \gets \{\}$ \Comment{Initialize Seed Documents}
\For{$c \in C$}
    \State $s \gets L(P)$ \Comment{Use LLM with DoPAMine prompt to generate seed documents}
    \State $S \gets S \cup \{s\}$ \Comment{Add seed document to set}
\EndFor
% \State $sV \gets \{\}$ \Comment{Initialize seed Vector set}
% \For{$s \in S$}
%     \State $sv \gets E.encode(s)$ \Comment{Encode seed document using Encoder Model}
%     \State $sV \gets sV \cup \{sv\}$ \Comment{Add vector to seed Vector Database}
% \EndFor

\State $N \gets \{\}$ \Comment{Initialize Nearest Neighbors}
\For{$s \in S$}
    \State $sv \gets E.encode(s)$ \Comment{Encode seed document using Encoder Model}
    \State $n \gets V.getNearestNeighbor(sv)$ \Comment{Retrieve nearest neighbors}
    \State $N \gets N \cup \{n\}$ \Comment{Add nearest neighbors to set}
\EndFor

\State $T \gets Labeling(N)$ \Comment{Add labels on nearest neighbors and use as classifier training data}

\Return $T$
\end{algorithmic}
\end{algorithm*}

Our approach revolves around the incremental pre-training of a large language model on labeled in-domain data, curated from web crawl data through an automated process - DoPAMine. The algorithm underpinning this approach is outlined in Algorithm \ref{algo1}. The overarching methodology encompasses the following principal components:

\subsection{Data processing and indexing}
Initially, we acquire a vast corpus of textual documents (D) from a diverse web crawl source- Common Crawl \cite{cc}. We process this data through a data-processing pipeline with the following steps: text extraction, document level and sub-document level de-duplication \cite{lee2022deduplicating} and gopher filtering \cite{rae2022scaling}. We then employ state-of-the-art encoder models to embed the entire corpus into a shared \(n-dim\) vector space. This embedding process maps the documents to dense vector representations, facilitating the learning of semantic similarities between documents across the corpus. We then index the vector representations of the corpus into a vector database \((V)\).

\subsection{Seed guided automated data mining}
This pivotal stage entails a multi-faceted process. We harness the capabilities of powerful generative language models, such as Claude 3 Sonnet \cite{claudev3} , to generate domain-specific synthetic seed documents. The approach entails carefully constructing chain-of-thought \cite{wei2023chainofthought} prompt templates, informed by heuristics that accounts for document types, personas, author demeanors, intended audiences, desired generation lengths, and other relevant parameters. We sample from different values for each of these parameters (table \ref{tab:dimensions}) to ensure randomness and diversity in seed generations. The choices are derived from an analysis of a representative sample of documents from the Common Crawl corpus. Subsequently, we feed these prompts to Claude 3 Sonnet \cite{claudev3} for synthetic seed generations. The resultant seed dataset comprises of independent and identically distributed (iid) domain-specific exemplars, providing a representative sample for one or more target domains. The prompt template employed for synthetic seed data generation is shown in figure \ref{fig:prompt}. This prompt instructs the LLM to follow a series of steps to finally generate meaningful domain specific synthetic seed data, as shown in table \ref{seed_examples}.

Next, we utilize the seed dataset to query for semantically similar documents within the indexed vector database $(V)$. For each seed document, we retrieve its nearest neighbors in the vector space, identifying texts in the corpus that exhibit semantic relatedness. We assess the similarity between documents in \(n-dim\) space using cosine similarity as our distance measure:

\begin{equation}
\mathrm{sim}(d_1, d_2) \approx \cos(\vec{d_1}, \vec{d_2}) = \frac{\vec{d_1} \cdot \vec{d_2}}{||\vec{d_1}|| \cdot ||\vec{d_2}||}
\end{equation}
where $\cos(\vec{d_1}, \vec{d_2})$ represents the cosine similarity between the vector representations of documents $d_1$ and $d_2$, a metric that captures their semantic relatedness. For each synthetic seed, we mine $k$ semantically similar documents, adhering to the following formulation:
\begin{equation}
\mathrm{NN}_k(d) = { d' \in D : \cos(\vec{d}, \vec{d'}) \geq t_{sim}}\label{eq:3}
\end{equation}

where $\mathrm{NN}_k(d)$ denotes the set of $k$ nearest neighbors for document $d$, $D$ represents the corpus, $\cos(\vec{d}, \vec{d'})$ is the cosine similarity between the vector representations of $d$ and $d'$, $t_\mathrm{sim}$ is a predefined similarity threshold. 
The importance of this step relies on the fact that, while LLM generated synthetic texts may appear coherent and topically relevant, they lack the nuance, factual reliability, and depth of research that characterizes \cite{Ji_2023, luo2024hallucination} authentic human-written documents. Mining semantically similar documents from the corpus $(D)$ using the synthetic seed data acts as a proxy to get real-world data for the domain, enabling a more authentic and reliable representation of the target industry.

\subsection{Classifier Modeling}
The quality of the mined documents \cite{Steck2024IsCO} in Eqn \ref{eq:3} is dependent on the similarity threshold $t_\mathrm{sim}$. We choose a high $t_\mathrm{sim}$ to extract only the most semantically similar documents per seed document from the indexed vector database $(V)$. This allows us to assign the mined document the domain label of the seed data. From a large corpus $D$ like the Common Crawl \cite{cc}, we can now mine and collate real-world domain-representative documents for all industries listed in table \ref{tab:dimensions}. It is possible that a mined document may belong to multiple industry domains. We address this by assigning multi-labels to mined documents by (a) finding mined documents that were retrieved as nearest neighbors for more than one domain, and (b) explicitly mining for documents with multiple domain relevance by generating synthetic seed data that are multiple domain oriented using more than one domain in the seed generation prompt. We then aggregate the retrieved documents for all target domains with the multi-labels to train a multi-label text classifier. The final trained model predicts domain labels for new texts from unlabeled documents. Note, one can choose to just work with one target domain but we present a broader implementation across different industry domains to showcase the versatility of the DoPAMine framework in cataloging a large unlabeled dataset.

\begin{table*}
\centering
\begin{tabular}{@{}|c p{15cm}}
\toprule
\textbf{Dimension} & \textbf{Choices} \\
\toprule
Doc type & Report, Blog post, News article, List of tweets, Press release, Email, Technical report, Textbook chapter, Research paper, Short story, Advertisement, Product proposal, Research proposal, Status update, Legal brief, Contract, Memo \\ 
\midrule
\multirow{5}{*}{Industry} & Media \& Entertainment, Financial Services, Sports, Public Sector, Education, Gaming, Retail, Software \& Internet, Travel \& Hospitality, Agriculture, Utilities, Healthcare \& Life sciences, Real Estate \& Construction, Manufacturing, Telecommunications, Automotive, Services, Consumer goods, Transportation \& Logistics, Law, Energy \\ 
\midrule
Length & Very long (>1k words), Long (>500 words), Short (<500 words) \\ 
\midrule
Demeanour & Professional, Angry, Bored, Informal, Sad, Excited, Confident, Exacting,  Poetic, Pedantic, Attentive to detail \\ \bottomrule
\end{tabular}
\caption{Seed generation prompt dimensions}
\label{tab:dimensions}
\end{table*}

\begin{figure}[!ht]
% \centering
\begin{tcolorbox}[colback=orange!10!white, % Background color
                  colframe=orange!10!white, % Frame color
                  width=\textwidth, % Width of the tcolorbox
                  arc=2mm, % Radius of the rounded corners
                  % auto outer arc,
                  ]
% \textbf{Seed data generation prompt}
% \hline
\begin{verbatim}

Write a {doc_type} about {industry} using the 
following steps. 

1. Generate a random topic from {industry} domain.    
2. Write a short premise for the {doc_type} about topic from the {industry}.
3. Write a short description of the author of the {doc_type}. The author 
should be a practicing member of the {industry} industry. 
4. Describe the {doc_type}'s audience.
5. Give the author's motive in writing the document for the audience. 
The author's demeanor is {demeanor}.
6. Write a {doc_type} about {industry} based on the topic generated,
premise, from the perspective and motive of the author targeting to the
audience.
The resulting document should be {length}. 

Your response should be in the following format.

    - TOPIC: 
    - PREMISE: 
    - AUTHOR: 
    - AUDIENCE: 
    - MOTIVE: 
    - DOCUMENT:
    
Response:   
\end{verbatim}
\end{tcolorbox}
\caption{Seed data generation prompt}
\label{fig:prompt}
\end{figure}

\subsection{Continued pre-training of LLM using mined domain data}
Finally, we use the classifier to successfully classify the remaining unlabeled documents in the large corpus $D$ making extraction of industry domain-specific data easy. This data was then directly used to continually pre-train and adapt a large language model to target domains in our ablation experiments. We incorporated the target domain data along with samples from previously seen training data distribution so as to avoid catastrophic forgetting \cite{Robins1995CatastrophicFR, rolnick2019experience}.

\section{Experiment Setup}
In this section we explain in detail our experiment setup for each components.

\subsection{Seed data generation} 
Leveraging the capabilities of Claude 3 Sonnet \cite{claudev3}, we generated 200 seed data samples per domain. This process spanned several industry domains (table \ref{tab:dimensions}), both with and without overlaps among them. For each domain, we generated diverse prompts by sampling across multiple dimensions: document types, author demeanors and lengths of generation. Complete list of prompt variation choices can be seen in table \ref{tab:dimensions}. We validated the diversity and realistic nature of the generated seed data by computing several lexical metrics and comparing them against real world web-crawled documents. Although we used retrieved real documents for Language Model Pretraining (LLM CPT), we wanted to ensure that the seed data used to retrieve those real documents were of realistic quality. Table \ref{tab:diversity} presents a comparison of the lexical metrics for the synthetic seed data and the real world web-crawled samples. The values across the different lexical metrics like Lexical Diversity \cite{10.1093/applin/amp024}, Readability \cite{6449109}, Hapax Legomena  \cite{Pierrehumbert2018OnHL}, and Lexical Richness  \cite{martínez2024bewarewordsevaluatinglexical} are quite similar for both the synthetic seed documents and real documents from the web crawl, indicating that the generated synthetic seed data closely resembles the real world web data in terms of diversity and lexical properties.

\begin{table}
\centering
\begin{tabular}{c|c|c}
\toprule
\multirow{2}{*}{\textbf{Metric}} & \multirow{1}{*}{\textbf{Real documents}} & \multirow{1}{*}{\textbf{Synthetic seed}} \\
\multirow{2}{*}{} & \multirow{1}{*}{\textbf{(CC)}} & \multirow{1}{*}{\textbf{documents}} \\
\midrule
Lexical Diversity \cite{10.1093/applin/amp024} & 0.63 & 0.62 \\
Flesch Readability \cite{6449109} & 10.08 & 9.8 \\
Hapax Legomena \cite{Pierrehumbert2018OnHL} & 0.564 & 0.559 \\
Lexical Richness \cite{martínez2024bewarewordsevaluatinglexical} & 92.16 & 137.64 \\
\bottomrule
\end{tabular}
\caption{Lexical metrics between real and synthetic documents}
\label{tab:diversity}
\end{table}

\subsection{Embedding generation}
We employed bge-large-en-v1.5 \cite{chen2024bge} encoder using sentence transformer \cite{reimers2019sentence} to convert the textual documents into vector representation. Since the embedding quality degrades when generated for very long documents, we pre-processed the common-crawl data to chunk the documents to maximum length of 2500 words respecting sentence boundaries. We also removed documents with fewer than 20 tokens to remove noise. The embeddings were then generated for the pre-processed documents.

\subsection{Nearest neighbor extraction}
We used OpenSearch \cite{opensearch} as a vector database choice to leverage its at-scale support for k-NN search at low-latency. To build the vector index, we used Non-Metric Space Library(NMSLIB) \cite{DBLP:conf/sisap/BoytsovN13} engine with Hierarchical Navigable Small World(HNSW) \cite{DBLP:journals/corr/MalkovY16} algorithm for knn search with support for cosine similarity distance metric. We experimented with various hyper-parameter settings optimizing for latency and recall to arrive at final settings of ef-construction:256, m:45 and ef-search:50. With respect to the index cluster configurations, to optimize for memory utilization, we kept the shard size at 15GB with 2 replicas of index to achieve latency of 45ms for 200 nearest neighbor lookup across shards in the index. After setting up the optimized index, we queried nearest neighbors for each synthetic seed generation. We choose a high $t_\mathrm{sim} = 0.85$ to extract only the most relevant nearest neighbors.

\subsection{Classifier training}
For the training of industry domain classifier model, we leveraged the NVIDIA A10G Tensor Core GPUs. Our training process encompassed a range of lightweight multilingual transformer candidates, such as m-distilbert \cite{sanh2020distilbert}, MiniLM \cite{wang2020minilm}, and others, as well as fastText \cite{bojanowski2017enriching} candidates. To ensure efficient application on terabytes of documents and to address latency concerns, we intentionally avoided employing large transformer models like XLMR \cite{conneau2020unsupervised} and Longformer \cite{beltagy2020longformer}. For training the models, we used 40k data samples and their translations in Latin, European and Asian languages to equip the model with multi-lingual capabilities.

\subsection{Ablation study}
For evaluating the usefulness of the in-domain data mined through DoPAMine, we conducted an ablation study by training 7B parameter LM candidates in the continual pre-training (CPT) setup. We employed computing clusters, each comprising 128 NVIDIA A100 GPUs, to train the LLMs with and without domain weighted data. We trained a base 7B parameter decoder only model and used it to initialize two candidate models to simulate continual pre-training scenario. This base model was trained with 300B web-crawl tokens without any domain filtering. Candidate models include a baseline and a target model that were trained using additional 100B crawl tokens with and without domain specific mined data, respectively. Both the baseline and the domain specific model are trained on a total fixed budget of 400B tokens (300B (base) + 100B (in CPT mode)) to keep the models comparable in terms of parametric knowledge. For the 100B tokens used for training the domain specific model we used a training data mixing ratio of 25\% (target industry domain data curated via DoPAMine) to 75\% (unlabled common crawl). The 75\% unlabeled common crawl training data is used to avoid catastrophic forgetting in the language model \cite{Robins1995CatastrophicFR, rolnick2019experience}.  Table \ref{tab:tdata} provides the training data token distribution across the different models trained in the ablation study. More details on the ablation study are explained in section \ref{ablation}.

\begin{table}[]
\centering
\begin{tabular}{l|l|l}
\toprule
\textbf{Model} & \textbf{Pretrain data} & \textbf{CPT data}                                                             \\ \hline
baseline       & 300B (CC)              & 100B (CC)                                                                     \\ \hline
DOP-healthcare & 300B (CC)              & \begin{tabular}[c]{@{}l@{}}25B (DoPAMine-healthcare),\\ 75B (CC)\end{tabular} \\ \hline
DOP-finance    & 300B (CC)              & \begin{tabular}[c]{@{}l@{}}25B (DoPAMine-finance),\\ 75B (CC)\end{tabular} \\
\bottomrule
\end{tabular}
\caption{Training data token distribution}
\label{tab:tdata}
\end{table}

\begin{table}[htbp]
\centering
\caption{Accuracy on healthcare evaluation datasets in zero- and five-shot evaluation settings}
\label{tab:hcresult}
\begin{tabular}{p{0.19\textwidth}|p{0.04\textwidth}p{0.04\textwidth}|p{0.04\textwidth}p{0.04\textwidth}} %{|l|c|c|c|c|}
\toprule
Dataset & \multicolumn{2}{c|}{baseline} & \multicolumn{2}{c}{DOP-healthcare} \\
\cline{2-5}
& zero & five & zero & five \\
\midrule
\textbf{MMLU}\cite{hendrycks2021measuring} &  &  &  &  \\
Anatomy & 15.0 & 23.5 & \textbf{18.3} & \textbf{27.2} \\
Clinical knowledge & 14.1 & 23.4 & \textbf{17.5} & \textbf{26.9} \\
College medicine & 16.1 & 28.0 & \textbf{19.4} & \textbf{30.3} \\
Human sexuality & 16.0 & 24.9 & \textbf{21.6} & \textbf{27.7} \\
Medical genetics & 13.0 & 24.0 & \textbf{16.2} & \textbf{25.1} \\
Pro medicine & 10.6 & 26.8 & \textbf{13.0} & \textbf{31.2} \\
Virology & 15.5 & 26.2 & \textbf{23.2} & \textbf{27.7} \\
 % &  &  & \textbf{23.2} & \textbf{27.7} \\
\\
\textbf{Mean} & 13.3 & 25.2 & \textbf{18.4} & \textbf{28.1} \\
\hdashline
\textbf{MedQA (USMLE)} \cite{jin2020disease} & 24.0  & 24.2  & \textbf{24.5}  & \textbf{25.0}  \\
\hdashline
\textbf{MedMCQA}\cite{pal2022medmcqa} & 22.5  & 25.1  & \textbf{23.1} & \textbf{26.0} \\
\hdashline
\textbf{PubMedQA}\cite{jin2019pubmedqa} & 41.2  & 55.3  & \textbf{54.8}  & 55.3 \\
\bottomrule
\end{tabular}
\end{table}

\begin{table}[!ht]
\centering
\caption{Accuracy on finance evaluation datasets in zero- and five-shot evaluation settings}
\label{tab:finresult}
\begin{tabular}{p{0.19\textwidth}|p{0.04\textwidth}p{0.04\textwidth}|p{0.04\textwidth}p{0.04\textwidth}} %{|l|c|c|c|c|}
\toprule
Dataset & \multicolumn{2}{c|}{baseline} & \multicolumn{2}{c}{DOP-finance} \\
\cline{2-5}
& zero & five & zero & five \\
\midrule
\textbf{FiQA-SA} \cite{inproceedings} & 47.7  & 48.7  & \textbf{53.3}  & \textbf{61.0}  \\
% \hline
\textbf{FPB} \cite{malo2013good} & 26.9  & 59.5  & \textbf{28.6} & 59.5 \\
% \hline
\textbf{Headlines}\cite{sinha2020impact} & 30.4   & 49.5  & \textbf{31.9}  & \textbf{57.3} \\
\bottomrule
\end{tabular}
\end{table}

Despite the subjectivity and potential biases of Llama 3 \cite{dubey2024llama3herdmodels} judgments, the high overall agreement percentage suggests that the industry domain classifier performed effectively and aligned well with the understanding of a state-of-the-art LLM.

\subsection{CPT Ablation Study} \label{ablation}
We perform an ablation study to evaluate the influence of incorporating domain-specific data mined via DoPAMine into a LLM's pre-training data. The underlying assumption is that incorporating domain-specific data would improve performance on downstream domain-specific tasks. Due to compute constraints, we limited our experiments to healthcare and finance domains as these areas are relatively specialized, and several benchmark datasets are available for testing downstream performance. DoPAMine was used to mine and collect healthcare and finance-specific data from web crawls. This  data was then injected at an increased proportion into the training dataset for two 7B parameter LLMs (DOP\_healthcare, DOP\_finance), while another 7B model (baseline) was trained without any domain weightage. 

\begin{figure*}[!thb]
  \centering
  \includegraphics[width=0.9\linewidth ]{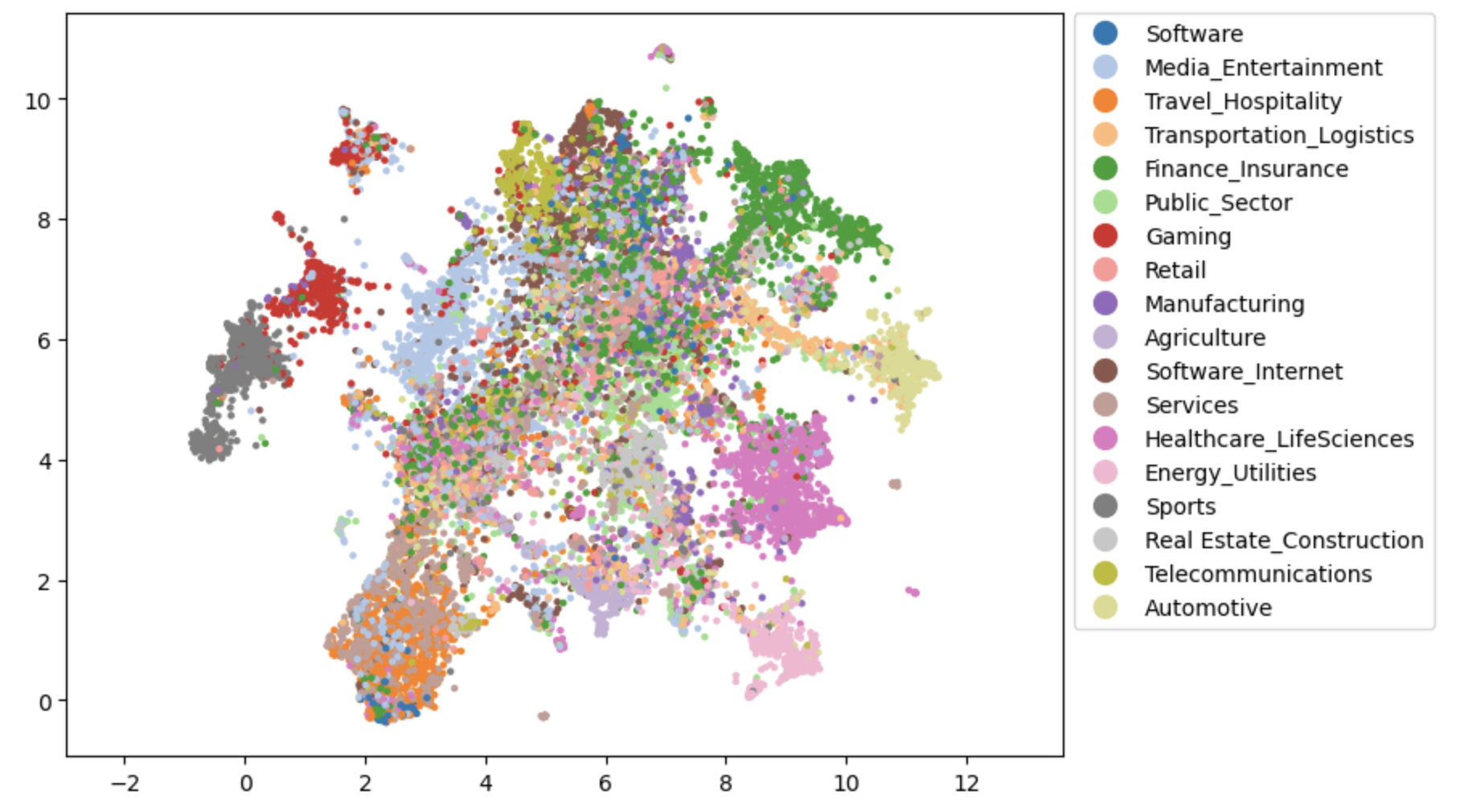}
  \caption{UMAP of mined in-domain data}
  \label{fig:umap}
\end{figure*}

All the models were initialized using a base 7B model that was trained using web-crawl documents(300B tokens) to simulate the continued pre-training (CPT) scenario. The performance of the DOP\_healthcare and baseline models was compared across several healthcare benchmark datastes similar to Med-PaLM 2 \cite{singhal2023expertlevel}. 

These included 7 MMLU \cite{hendrycks2021measuring} tasks related to healthcare domains (anatomy, clinical knowledge, college medicine, human sexuality, medical genetics, professional medicine, and virology), MedQA (USMLE) \cite{jin2020disease}, PubMedQA \cite{jin2019pubmedqa} and MedMCQA \cite{pal2022medmcqa}. Similarly, the DOP\_finance and baseline models were compared across several finance datasets akin to \cite{wu2023bloomberggpt, cheng2024adapting} comprising two sentiment analysis datasets (FiQA-SA \cite{inproceedings} and FPB \cite{malo2013good}) and one binary classification dataset (Headlines \cite{sinha2020impact}) of financial news headlines. Details about the datasets can be seen in Appendix \ref{datasets}. Both Zero-shot and Five-shot evaluation scenarios were covered under these tasks as the models were pre-trained only but not instruction fine-tuned. The results demonstrated that the DOP\_healthcare and DOP\_finance models, trained with proportionally more healthcare and finance data curated by DoPAMine, showed improved performance across all the domain-specific evaluation tasks and evaluation scenarios when compared to the baseline model, as shown in table \ref{tab:hcresult} and table \ref{tab:finresult}.\par
This ablation study suggests that incorporating domain-specific data into LLM training, mined using DoPAMine approach, can effectively enhance the model's performance on downstream tasks within that domain. DoPAMine plays a crucial role in identifying and selecting relevant and real in-domain data, leading to improved specialization and capability of the trained language model.

\subsection{Visualization of Mined Domain Data}
The Uniform Manifold Approximation and Projection (UMAP) \cite{mcinnes2020umap} technique was leveraged to visualize and gain insights into the domain-specific data mined by the proposed framework. The UMAP \cite{mcinnes2020umap} visualization in figure \ref{fig:umap} reveals distinct clusters for different industry domains like healthcare, finance, and technology, validating the framework's effectiveness in identifying and separating domain-specific documents. Interestingly, overlapping regions between certain domain clusters suggest shared characteristics or content, enabling multi-label assignment and cross-domain applications. A dense central region represents documents with general content relevant across multiple domains, potentially enhancing the language model's performance on tasks requiring broad domain knowledge or transfer learning capabilities. Overall, the UMAP visualization not only validates the data mining approach but also provides valuable insights into the relationships and interconnections between domains, informing the development of more targeted, specialized, or multi-task language models.

\begin{table*}[htbp]
\centering
\caption{Seed Generation Example: Healthcare and Life Sciences domain}
\begin{tabular}{p{\textwidth}}
\toprule
\textbf{Seed Generation Example: Healthcare and Life Sciences domain}  \\
\midrule
\textbf{Prompt:}
\begin{Verbatim}[commandchars=+\[\]]
Prompt: Write a +underline[product proposal] about +underline[Healthcare & Life Sciences] 
using the following steps.
1. Generate a random topic from Healthcare & Life Sciences domain.
2. Write a short premise for the +underline[product proposal] about the 
topic from +underline[Healthcare & Life Sciences] 
3. Write a short description of the author of the +underline[product proposal].
The author should be a practicing member of the +underline[Healthcare & Life Sciences] industry.
4. Describe the +underline[product proposal]'s audience.
5. Give the author's motive in writing the document for the audience. The author's 
demeanor is +underline[professional].
6. Write a +underline[product proposal] about +underline[Healthcare & Life Sciences] based on the topic 
generated, premise from the perspective and motive of the author targeting to the 
audience.The resulting document should be +underline[short (less than 500 words)]. 
Your response should be in the following format.
        - TOPIC: 
        - PREMISE: 
        - AUTHOR:
        - AUDIENCE:
        - MOTIVE: 
        - DOCUMENT:
\end{Verbatim}
% \midrule
\textbf{Generation:}
\begin{Verbatim}[commandchars=+\[\]]
- TOPIC: Wearable Health Monitoring Devices for Elderly Care
- PREMISE: The product proposal aims to introduce a new line of wearable health 
monitoring devices specifically designed for elderly individuals,...<truncated>
- AUTHOR: Dr. Emily Thompson is a renowned geriatrician with over 20 years of 
experience in the healthcare industry. She currently serves as the Chief 
Medical Officer <truncated>
- AUDIENCE: The product proposal is targeted towards healthcare providers,
caregivers,and decision-makers in the elderly care sector,... <truncated>
- MOTIVE: Dr. Emily Thompson's primary motive in writing this product 
proposal is toaddress the growing demand for innovative healthcare 
solutions... <truncated> 
- DOCUMENT: Dear Healthcare Professionals and Decision-Makers,
As our population continues to age, the demand for innovative healthcare 
solutions tailored to the unique needs of elderly individuals has never
been greater. In response to this pressing need,I am excited to 
introduce... <truncated>
\end{Verbatim}
\\
\bottomrule
\end{tabular}

\label{seed_examples}
\end{table*}

\section{Related Works}
The task of curating domain-specific training data for large language models has garnered significant attention in recent years, as researchers and practitioners recognize the importance of tailoring these models to specific domains and applications \cite{wang2024role, saito2024unsupervised, luo2023empirical}. Our work builds upon and contributes to several lines of research in natural language processing \cite{bespalov-etal-2023-towards, together2023redpajama, longpre2023pretrainersguidetrainingdata, brand2022text, yang2022re3generatinglongerstories}, data mining, and machine learning. In the recent work on TinyStories \cite{eldan2023tinystories}  (a high quality dataset synthetically generated to teach English to neural networks), showcased the impact of high quality data on LLM evaluation metrics, potentially allowing to match the performance of large-scale models with much smaller training/models. Furthermore, recently SOTA results were achieved during training of "phi-1" \cite{gunasekar2023textbooks}, a new large language model for code, with significantly smaller size(1.3B) trained using synthetically generated textbooks and exercises dataset with GPT-3.5 (1B tokens). Synthetically curated "textbook" like data was used to train the model providing clear, self-contained, instructive, and balanced examples of coding concepts and skills to enhance the learning process compared to traditional web data. The authors further trained phi-1.5 \cite{li2023textbooks} version of model focusing on common sense reasoning in natural language, to again strongly establish that high-quality synthetic datasets can lead to models that outperform SOTA models 5x larger in size on natural language tasks. Developing high quality dataset for specific domain presents key challenges namely, ensuring richness and distinctiveness in the dataset. Diversity entails a broad coverage, scenarios spanning varied levels of difficulty, complexity, and stylistic nuances. It serves multiple purposes, such as mitigating overfitting risks and bolstering the model's adaptability to novel tasks. However, attaining such diversity poses challenges, mere prompting for coding textbook-like content or exercises, albeit with some tweaks in instructions or parameters, is prone to yielding a monotonous dataset, wherein identical concepts and solutions recur with minor alterations \cite{ding2023enhancing, wei2023magicoder, toshniwal2024openmathinstruct1, cosmopedia}. In this paper, we dive into strategies to enforce diversity and randomness in data generation process to mitigate the above risks, and propose a framework to mine high-quality domain specific datasets that can be leveraged for LLM training.

\section{Limitations and Future Works}
While extensive experiments were conducted to validate the usefulness of DoPAMine, certain limitations exist, along with opportunities for future work. We conducted the ablation experiments for the domain specific model with a mixing ratio setting of 25\% curated in-domain data identified by DoPAMine and 75\% samples from unlabeled common crawl data to avoid catastrophic forgetting. Ideally one could vary this mixing ratio to find an upper bound of curated in-domain data which does not lead to catastrophic forgetting and improves downstream task performance. This exercise is extraneous to the DoPAMine methodology and hence we dont spend compute resources to find the optimal mixing ratio. Furthermore, the experiments utilized Claude 3 Sonnet for synthetic data generation and LLama3 as the judge model, as they were the most advanced accessible models for us at the time. However, employing the latest state-of-the-art models, such as Claude 3.5 Sonnet or Claude 3 Opus, could potentially enhance the quality of synthetic data and improve overall performance. Additionally, for classifier training, real documents were used post nearest-neighbor search, but an alternative approach could involve training the classifier directly with synthetic data and using it for mining real data. Moreover, in the experiments, real documents mined using DoPAMine were utilized for domain adaptation training, but combining some synthetic data with this curated data might improve performance, as evidenced by recent works \cite{li2023textbooks}.

\section{Conclusion}
DoPAMine is an automated and scalable framework that leverages large language models and carefully crafted prompts to generate diverse, representative seed data tailored to specific industry domains. This seed data guides the retrieval of semantically similar real-world documents from large corpora like web crawls, curating domain-specific datasets for pre-training language models. Extensive experimentation demonstrates DoPAMine's effectiveness in improving language model performance on downstream domain tasks by incorporating the curated data during continued pre-training. The framework offers several advantages over existing methods, enabling scalable and precise control over target domains, including low-resource ones, while mitigating issues with relying solely on LLM generated synthetic data. By facilitating the efficient development of domain-tailored language models across fields like healthcare, finance, etc, DoPAMine represents a significant step towards providing more accurate and specialized natural language processing capabilities, unlocking new possibilities and driving advancements and LLM adoption across various industries.

% \nocite{*}
% \printbibliography
\bibliographystyle{plain}

% \bibliography{references}{}

\begin{thebibliography}{10}

\bibitem{claudev3}
Anthropic.
\newblock The claude 3 model family: Opus, sonnet, haiku.

\bibitem{arannil-etal-2023-adeqa}
Vinayak Arannil, Tomal Deb, and Atanu Roy.
\newblock {ADEQA}: A question answer based approach for joint {ADE}-suspect extraction using sequence-to-sequence transformers.
\newblock In Dina Demner-fushman, Sophia Ananiadou, and Kevin Cohen, editors, {\em The 22nd Workshop on Biomedical Natural Language Processing and BioNLP Shared Tasks}, pages 206--214, Toronto, Canada, July 2023. Association for Computational Linguistics.

\bibitem{beltagy2020longformer}
Iz~Beltagy, Matthew~E. Peters, and Arman Cohan.
\newblock Longformer: The long-document transformer, 2020.

\bibitem{bespalov-etal-2023-towards}
Dmitriy Bespalov, Sourav Bhabesh, Yi~Xiang, Liutong Zhou, and Yanjun Qi.
\newblock Towards building a robust toxicity predictor.
\newblock In Sunayana Sitaram, Beata Beigman~Klebanov, and Jason~D Williams, editors, {\em Proceedings of the 61st Annual Meeting of the Association for Computational Linguistics (Volume 5: Industry Track)}, pages 581--598, Toronto, Canada, July 2023. Association for Computational Linguistics.

\bibitem{bojanowski2017enriching}
Piotr Bojanowski, Edouard Grave, Armand Joulin, and Tomas Mikolov.
\newblock Enriching word vectors with subword information, 2017.

\bibitem{pmlr-v162-borgeaud22a}
Sebastian Borgeaud, Arthur Mensch, Jordan Hoffmann, Trevor Cai, Eliza Rutherford, Katie Millican, George~Bm Van Den~Driessche, Jean-Baptiste Lespiau, Bogdan Damoc, Aidan Clark, Diego De~Las~Casas, Aurelia Guy, Jacob Menick, Roman Ring, Tom Hennigan, Saffron Huang, Loren Maggiore, Chris Jones, Albin Cassirer, Andy Brock, Michela Paganini, Geoffrey Irving, Oriol Vinyals, Simon Osindero, Karen Simonyan, Jack Rae, Erich Elsen, and Laurent Sifre.
\newblock Improving language models by retrieving from trillions of tokens.
\newblock In Kamalika Chaudhuri, Stefanie Jegelka, Le~Song, Csaba Szepesvari, Gang Niu, and Sivan Sabato, editors, {\em Proceedings of the 39th International Conference on Machine Learning}, volume 162 of {\em Proceedings of Machine Learning Research}, pages 2206--2240. PMLR, 17--23 Jul 2022.

\bibitem{DBLP:conf/sisap/BoytsovN13}
Leonid Boytsov and Bilegsaikhan Naidan.
\newblock Engineering efficient and effective non-metric space library.
\newblock In Nieves~R. Brisaboa, Oscar Pedreira, and Pavel Zezula, editors, {\em Similarity Search and Applications - 6th International Conference, {SISAP} 2013, {A} Coru{\~{n}}a, Spain, October 2-4, 2013, Proceedings}, volume 8199 of {\em Lecture Notes in Computer Science}, pages 280--293. Springer, 2013.

\bibitem{brand2022text}
Ryan Brand, Sia Gholami, Daniel Horowitz, Liutong Zhou, and Sourav Bhabesh.
\newblock Text classification for online conversations with machine learning on aws.
\newblock {\em AWS Machine Learning Blog}, 2022.

\bibitem{brown2020language}
Tom Brown, Benjamin Mann, Nick Ryder, Melanie Subbiah, Jared~D Kaplan, Prafulla Dhariwal, Arvind Neelakantan, Pranav Shyam, Girish Sastry, Amanda Askell, et~al.
\newblock Language models are few-shot learners.
\newblock {\em Advances in neural information processing systems}, 33:1877--1901, 2020.

\bibitem{chen2024bge}
Jianlv Chen, Shitao Xiao, Peitian Zhang, Kun Luo, Defu Lian, and Zheng Liu.
\newblock Bge m3-embedding: Multi-lingual, multi-functionality, multi-granularity text embeddings through self-knowledge distillation, 2024.

\bibitem{cheng2024adapting}
Daixuan Cheng, Shaohan Huang, and Furu Wei.
\newblock Adapting large language models via reading comprehension, 2024.

\bibitem{chowdhery2023palm}
Aakanksha Chowdhery, Sharan Narang, Jacob Devlin, Maarten Bosma, Gaurav Mishra, Adam Roberts, Paul Barham, Hyung~Won Chung, Charles Sutton, Sebastian Gehrmann, et~al.
\newblock Palm: Scaling language modeling with pathways.
\newblock {\em Journal of Machine Learning Research}, 24(240):1--113, 2023.

\bibitem{together2023redpajama}
Together Computer.
\newblock Redpajama: an open dataset for training large language models, 2023.

\bibitem{conneau2020unsupervised}
Alexis Conneau, Kartikay Khandelwal, Naman Goyal, Vishrav Chaudhary, Guillaume Wenzek, Francisco Guzmán, Edouard Grave, Myle Ott, Luke Zettlemoyer, and Veselin Stoyanov.
\newblock Unsupervised cross-lingual representation learning at scale, 2020.

\bibitem{cosmopedia}
cosmopedia.
\newblock cosmopedia.

\bibitem{cc}
Common Crawl.
\newblock Common crawl.

\bibitem{ding2023enhancing}
Ning Ding, Yulin Chen, Bokai Xu, Yujia Qin, Zhi Zheng, Shengding Hu, Zhiyuan Liu, Maosong Sun, and Bowen Zhou.
\newblock Enhancing chat language models by scaling high-quality instructional conversations, 2023.

\bibitem{dodge2021documenting}
Jesse Dodge, Maarten Sap, Ana Marasovi{\'c}, William Agnew, Gabriel Ilharco, Dirk Groeneveld, Margaret Mitchell, and Matt Gardner.
\newblock Documenting large webtext corpora: A case study on the colossal clean crawled corpus.
\newblock {\em arXiv preprint arXiv:2104.08758}, 2021.

\bibitem{dubey2024llama3herdmodels}
Abhimanyu Dubey, Abhinav Jauhri, Abhinav Pandey, Abhishek Kadian, Ahmad Al-Dahle, Aiesha Letman, Akhil Mathur, Alan Schelten, Amy Yang, Angela Fan, Anirudh Goyal, Anthony Hartshorn, Aobo Yang, Archi Mitra, Archie Sravankumar, Artem Korenev, Arthur Hinsvark, Arun Rao, Aston Zhang, Aurelien Rodriguez, Austen Gregerson, Ava Spataru, Baptiste Roziere, Bethany Biron, Binh Tang, Bobbie Chern, Charlotte Caucheteux, Chaya Nayak, Chloe Bi, Chris Marra, Chris McConnell, Christian Keller, Christophe Touret, Chunyang Wu, Corinne Wong, Cristian~Canton Ferrer, Cyrus Nikolaidis, Damien Allonsius, Daniel Song, Danielle Pintz, Danny Livshits, David Esiobu, Dhruv Choudhary, Dhruv Mahajan, Diego Garcia-Olano, Diego Perino, Dieuwke Hupkes, Egor Lakomkin, Ehab AlBadawy, Elina Lobanova, Emily Dinan, Eric~Michael Smith, Filip Radenovic, Frank Zhang, Gabriel Synnaeve, Gabrielle Lee, Georgia~Lewis Anderson, Graeme Nail, Gregoire Mialon, Guan Pang, Guillem Cucurell, Hailey Nguyen, Hannah Korevaar, Hu~Xu, Hugo Touvron, Iliyan Zarov,
  Imanol~Arrieta Ibarra, Isabel Kloumann, Ishan Misra, Ivan Evtimov, Jade Copet, Jaewon Lee, Jan Geffert, Jana Vranes, Jason Park, Jay Mahadeokar, Jeet Shah, Jelmer van~der Linde, Jennifer Billock, Jenny Hong, Jenya Lee, Jeremy Fu, Jianfeng Chi, Jianyu Huang, Jiawen Liu, Jie Wang, Jiecao Yu, Joanna Bitton, Joe Spisak, Jongsoo Park, Joseph Rocca, Joshua Johnstun, Joshua Saxe, Junteng Jia, Kalyan~Vasuden Alwala, Kartikeya Upasani, Kate Plawiak, Ke~Li, Kenneth Heafield, Kevin Stone, Khalid El-Arini, Krithika Iyer, Kshitiz Malik, Kuenley Chiu, Kunal Bhalla, Lauren Rantala-Yeary, Laurens van~der Maaten, Lawrence Chen, Liang Tan, Liz Jenkins, Louis Martin, Lovish Madaan, Lubo Malo, Lukas Blecher, Lukas Landzaat, Luke de~Oliveira, Madeline Muzzi, Mahesh Pasupuleti, Mannat Singh, Manohar Paluri, Marcin Kardas, Mathew Oldham, Mathieu Rita, Maya Pavlova, Melanie Kambadur, Mike Lewis, Min Si, Mitesh~Kumar Singh, Mona Hassan, Naman Goyal, Narjes Torabi, Nikolay Bashlykov, Nikolay Bogoychev, Niladri Chatterji, Olivier
  Duchenne, Onur Çelebi, Patrick Alrassy, Pengchuan Zhang, Pengwei Li, Petar Vasic, Peter Weng, Prajjwal Bhargava, Pratik Dubal, Praveen Krishnan, Punit~Singh Koura, Puxin Xu, Qing He, Qingxiao Dong, Ragavan Srinivasan, Raj Ganapathy, Ramon Calderer, Ricardo~Silveira Cabral, Robert Stojnic, Roberta Raileanu, Rohit Girdhar, Rohit Patel, Romain Sauvestre, Ronnie Polidoro, Roshan Sumbaly, Ross Taylor, Ruan Silva, Rui Hou, Rui Wang, Saghar Hosseini, Sahana Chennabasappa, Sanjay Singh, Sean Bell, Seohyun~Sonia Kim, Sergey Edunov, Shaoliang Nie, Sharan Narang, Sharath Raparthy, Sheng Shen, Shengye Wan, Shruti Bhosale, Shun Zhang, Simon Vandenhende, Soumya Batra, Spencer Whitman, Sten Sootla, Stephane Collot, Suchin Gururangan, Sydney Borodinsky, Tamar Herman, Tara Fowler, Tarek Sheasha, Thomas Georgiou, Thomas Scialom, Tobias Speckbacher, Todor Mihaylov, Tong Xiao, Ujjwal Karn, Vedanuj Goswami, Vibhor Gupta, Vignesh Ramanathan, Viktor Kerkez, Vincent Gonguet, Virginie Do, Vish Vogeti, Vladan Petrovic, Weiwei Chu,
  Wenhan Xiong, Wenyin Fu, Whitney Meers, Xavier Martinet, Xiaodong Wang, Xiaoqing~Ellen Tan, Xinfeng Xie, Xuchao Jia, Xuewei Wang, Yaelle Goldschlag, Yashesh Gaur, Yasmine Babaei, Yi~Wen, Yiwen Song, Yuchen Zhang, Yue Li, Yuning Mao, Zacharie~Delpierre Coudert, Zheng Yan, Zhengxing Chen, Zoe Papakipos, Aaditya Singh, Aaron Grattafiori, Abha Jain, Adam Kelsey, Adam Shajnfeld, Adithya Gangidi, Adolfo Victoria, Ahuva Goldstand, Ajay Menon, Ajay Sharma, Alex Boesenberg, Alex Vaughan, Alexei Baevski, Allie Feinstein, Amanda Kallet, Amit Sangani, Anam Yunus, Andrei Lupu, Andres Alvarado, Andrew Caples, Andrew Gu, Andrew Ho, Andrew Poulton, Andrew Ryan, Ankit Ramchandani, Annie Franco, Aparajita Saraf, Arkabandhu Chowdhury, Ashley Gabriel, Ashwin Bharambe, Assaf Eisenman, Azadeh Yazdan, Beau James, Ben Maurer, Benjamin Leonhardi, Bernie Huang, Beth Loyd, Beto~De Paola, Bhargavi Paranjape, Bing Liu, Bo~Wu, Boyu Ni, Braden Hancock, Bram Wasti, Brandon Spence, Brani Stojkovic, Brian Gamido, Britt Montalvo, Carl
  Parker, Carly Burton, Catalina Mejia, Changhan Wang, Changkyu Kim, Chao Zhou, Chester Hu, Ching-Hsiang Chu, Chris Cai, Chris Tindal, Christoph Feichtenhofer, Damon Civin, Dana Beaty, Daniel Kreymer, Daniel Li, Danny Wyatt, David Adkins, David Xu, Davide Testuggine, Delia David, Devi Parikh, Diana Liskovich, Didem Foss, Dingkang Wang, Duc Le, Dustin Holland, Edward Dowling, Eissa Jamil, Elaine Montgomery, Eleonora Presani, Emily Hahn, Emily Wood, Erik Brinkman, Esteban Arcaute, Evan Dunbar, Evan Smothers, Fei Sun, Felix Kreuk, Feng Tian, Firat Ozgenel, Francesco Caggioni, Francisco Guzmán, Frank Kanayet, Frank Seide, Gabriela~Medina Florez, Gabriella Schwarz, Gada Badeer, Georgia Swee, Gil Halpern, Govind Thattai, Grant Herman, Grigory Sizov, Guangyi, Zhang, Guna Lakshminarayanan, Hamid Shojanazeri, Han Zou, Hannah Wang, Hanwen Zha, Haroun Habeeb, Harrison Rudolph, Helen Suk, Henry Aspegren, Hunter Goldman, Igor Molybog, Igor Tufanov, Irina-Elena Veliche, Itai Gat, Jake Weissman, James Geboski, James Kohli,
  Japhet Asher, Jean-Baptiste Gaya, Jeff Marcus, Jeff Tang, Jennifer Chan, Jenny Zhen, Jeremy Reizenstein, Jeremy Teboul, Jessica Zhong, Jian Jin, Jingyi Yang, Joe Cummings, Jon Carvill, Jon Shepard, Jonathan McPhie, Jonathan Torres, Josh Ginsburg, Junjie Wang, Kai Wu, Kam~Hou U, Karan Saxena, Karthik Prasad, Kartikay Khandelwal, Katayoun Zand, Kathy Matosich, Kaushik Veeraraghavan, Kelly Michelena, Keqian Li, Kun Huang, Kunal Chawla, Kushal Lakhotia, Kyle Huang, Lailin Chen, Lakshya Garg, Lavender A, Leandro Silva, Lee Bell, Lei Zhang, Liangpeng Guo, Licheng Yu, Liron Moshkovich, Luca Wehrstedt, Madian Khabsa, Manav Avalani, Manish Bhatt, Maria Tsimpoukelli, Martynas Mankus, Matan Hasson, Matthew Lennie, Matthias Reso, Maxim Groshev, Maxim Naumov, Maya Lathi, Meghan Keneally, Michael~L. Seltzer, Michal Valko, Michelle Restrepo, Mihir Patel, Mik Vyatskov, Mikayel Samvelyan, Mike Clark, Mike Macey, Mike Wang, Miquel~Jubert Hermoso, Mo~Metanat, Mohammad Rastegari, Munish Bansal, Nandhini Santhanam, Natascha
  Parks, Natasha White, Navyata Bawa, Nayan Singhal, Nick Egebo, Nicolas Usunier, Nikolay~Pavlovich Laptev, Ning Dong, Ning Zhang, Norman Cheng, Oleg Chernoguz, Olivia Hart, Omkar Salpekar, Ozlem Kalinli, Parkin Kent, Parth Parekh, Paul Saab, Pavan Balaji, Pedro Rittner, Philip Bontrager, Pierre Roux, Piotr Dollar, Polina Zvyagina, Prashant Ratanchandani, Pritish Yuvraj, Qian Liang, Rachad Alao, Rachel Rodriguez, Rafi Ayub, Raghotham Murthy, Raghu Nayani, Rahul Mitra, Raymond Li, Rebekkah Hogan, Robin Battey, Rocky Wang, Rohan Maheswari, Russ Howes, Ruty Rinott, Sai~Jayesh Bondu, Samyak Datta, Sara Chugh, Sara Hunt, Sargun Dhillon, Sasha Sidorov, Satadru Pan, Saurabh Verma, Seiji Yamamoto, Sharadh Ramaswamy, Shaun Lindsay, Shaun Lindsay, Sheng Feng, Shenghao Lin, Shengxin~Cindy Zha, Shiva Shankar, Shuqiang Zhang, Shuqiang Zhang, Sinong Wang, Sneha Agarwal, Soji Sajuyigbe, Soumith Chintala, Stephanie Max, Stephen Chen, Steve Kehoe, Steve Satterfield, Sudarshan Govindaprasad, Sumit Gupta, Sungmin Cho, Sunny
  Virk, Suraj Subramanian, Sy~Choudhury, Sydney Goldman, Tal Remez, Tamar Glaser, Tamara Best, Thilo Kohler, Thomas Robinson, Tianhe Li, Tianjun Zhang, Tim Matthews, Timothy Chou, Tzook Shaked, Varun Vontimitta, Victoria Ajayi, Victoria Montanez, Vijai Mohan, Vinay~Satish Kumar, Vishal Mangla, Vlad Ionescu, Vlad Poenaru, Vlad~Tiberiu Mihailescu, Vladimir Ivanov, Wei Li, Wenchen Wang, Wenwen Jiang, Wes Bouaziz, Will Constable, Xiaocheng Tang, Xiaofang Wang, Xiaojian Wu, Xiaolan Wang, Xide Xia, Xilun Wu, Xinbo Gao, Yanjun Chen, Ye~Hu, Ye~Jia, Ye~Qi, Yenda Li, Yilin Zhang, Ying Zhang, Yossi Adi, Youngjin Nam, Yu, Wang, Yuchen Hao, Yundi Qian, Yuzi He, Zach Rait, Zachary DeVito, Zef Rosnbrick, Zhaoduo Wen, Zhenyu Yang, and Zhiwei Zhao.
\newblock The llama 3 herd of models, 2024.

\bibitem{eldan2023tinystories}
Ronen Eldan and Yuanzhi Li.
\newblock Tinystories: How small can language models be and still speak coherent english?, 2023.

\bibitem{pmlr-v202-fu23d}
Yao Fu, Hao Peng, Litu Ou, Ashish Sabharwal, and Tushar Khot.
\newblock Specializing smaller language models towards multi-step reasoning.
\newblock In Andreas Krause, Emma Brunskill, Kyunghyun Cho, Barbara Engelhardt, Sivan Sabato, and Jonathan Scarlett, editors, {\em Proceedings of the 40th International Conference on Machine Learning}, volume 202 of {\em Proceedings of Machine Learning Research}, pages 10421--10430. PMLR, 23--29 Jul 2023.

\bibitem{gunasekar2023textbooks}
Suriya Gunasekar, Yi~Zhang, Jyoti Aneja, Caio C{\'e}sar~Teodoro Mendes, Allie Del~Giorno, Sivakanth Gopi, Mojan Javaheripi, Piero Kauffmann, Gustavo de~Rosa, Olli Saarikivi, et~al.
\newblock Textbooks are all you need.
\newblock {\em arXiv preprint arXiv:2306.11644}, 2023.

\bibitem{hendrycks2021measuring}
Dan Hendrycks, Collin Burns, Steven Basart, Andy Zou, Mantas Mazeika, Dawn Song, and Jacob Steinhardt.
\newblock Measuring massive multitask language understanding, 2021.

\bibitem{hoffmann2022training}
Jordan Hoffmann, Sebastian Borgeaud, Arthur Mensch, Elena Buchatskaya, Trevor Cai, Eliza Rutherford, Diego de~Las Casas, Lisa~Anne Hendricks, Johannes Welbl, Aidan Clark, et~al.
\newblock Training compute-optimal large language models.
\newblock {\em arXiv preprint arXiv:2203.15556}, 2022.

\bibitem{huang2024empirical}
Hui Huang, Yingqi Qu, Jing Liu, Muyun Yang, and Tiejun Zhao.
\newblock An empirical study of llm-as-a-judge for llm evaluation: Fine-tuned judge models are task-specific classifiers.
\newblock {\em arXiv preprint arXiv:2403.02839}, 2024.

\bibitem{Ji_2023}
Ziwei Ji, Nayeon Lee, Rita Frieske, Tiezheng Yu, Dan Su, Yan Xu, Etsuko Ishii, Ye~Jin Bang, Andrea Madotto, and Pascale Fung.
\newblock Survey of hallucination in natural language generation.
\newblock {\em ACM Computing Surveys}, 55(12):1–38, March 2023.

\bibitem{jin2020disease}
Di~Jin, Eileen Pan, Nassim Oufattole, Wei-Hung Weng, Hanyi Fang, and Peter Szolovits.
\newblock What disease does this patient have? a large-scale open domain question answering dataset from medical exams, 2020.

\bibitem{jin2024comprehensive}
Hanlei Jin, Yang Zhang, Dan Meng, Jun Wang, and Jinghua Tan.
\newblock A comprehensive survey on process-oriented automatic text summarization with exploration of llm-based methods, 2024.

\bibitem{jin2019pubmedqa}
Qiao Jin, Bhuwan Dhingra, Zhengping Liu, William~W. Cohen, and Xinghua Lu.
\newblock Pubmedqa: A dataset for biomedical research question answering, 2019.

\bibitem{kim2023sure}
Jaehyung Kim, Jaehyun Nam, Sangwoo Mo, Jongjin Park, Sang-Woo Lee, Minjoon Seo, Jung-Woo Ha, and Jinwoo Shin.
\newblock Sure: Improving open-domain question answering of llms via summarized retrieval.
\newblock In {\em The Twelfth International Conference on Learning Representations}, 2023.

\bibitem{lee2022deduplicating}
Katherine Lee, Daphne Ippolito, Andrew Nystrom, Chiyuan Zhang, Douglas Eck, Chris Callison-Burch, and Nicholas Carlini.
\newblock Deduplicating training data makes language models better, 2022.

\bibitem{NEURIPS2020_6b493230}
Patrick Lewis, Ethan Perez, Aleksandra Piktus, Fabio Petroni, Vladimir Karpukhin, Naman Goyal, Heinrich K\"{u}ttler, Mike Lewis, Wen-tau Yih, Tim Rockt\"{a}schel, Sebastian Riedel, and Douwe Kiela.
\newblock Retrieval-augmented generation for knowledge-intensive nlp tasks.
\newblock In H.~Larochelle, M.~Ranzato, R.~Hadsell, M.F. Balcan, and H.~Lin, editors, {\em Advances in Neural Information Processing Systems}, volume~33, pages 9459--9474. Curran Associates, Inc., 2020.

\bibitem{li2023textbooks}
Yuanzhi Li, Sébastien Bubeck, Ronen Eldan, Allie~Del Giorno, Suriya Gunasekar, and Yin~Tat Lee.
\newblock Textbooks are all you need ii: phi-1.5 technical report, 2023.

\bibitem{longpre2023pretrainersguidetrainingdata}
Shayne Longpre, Gregory Yauney, Emily Reif, Katherine Lee, Adam Roberts, Barret Zoph, Denny Zhou, Jason Wei, Kevin Robinson, David Mimno, and Daphne Ippolito.
\newblock A pretrainer's guide to training data: Measuring the effects of data age, domain coverage, quality, \& toxicity, 2023.

\bibitem{luo2024hallucination}
Junliang Luo, Tianyu Li, Di~Wu, Michael Jenkin, Steve Liu, and Gregory Dudek.
\newblock Hallucination detection and hallucination mitigation: An investigation, 2024.

\bibitem{luo2023empirical}
Yun Luo, Zhen Yang, Fandong Meng, Yafu Li, Jie Zhou, and Yue Zhang.
\newblock An empirical study of catastrophic forgetting in large language models during continual fine-tuning.
\newblock {\em arXiv preprint arXiv:2308.08747}, 2023.

\bibitem{inproceedings}
Macedo Maia, Siegfried Handschuh, Andre Freitas, Brian Davis, Ross McDermott, Manel Zarrouk, and Alexandra Balahur.
\newblock Www'18 open challenge: Financial opinion mining and question answering.
\newblock pages 1941--1942, 04 2018.

\bibitem{DBLP:journals/corr/MalkovY16}
Yury~A. Malkov and Dmitry~A. Yashunin.
\newblock Efficient and robust approximate nearest neighbor search using hierarchical navigable small world graphs.
\newblock {\em CoRR}, abs/1603.09320, 2016.

\bibitem{malo2013good}
Pekka Malo, Ankur Sinha, Pyry Takala, Pekka Korhonen, and Jyrki Wallenius.
\newblock Good debt or bad debt: Detecting semantic orientations in economic texts, 2013.

\bibitem{martínez2024bewarewordsevaluatinglexical}
Gonzalo Martínez, José~Alberto Hernández, Javier Conde, Pedro Reviriego, and Elena Merino.
\newblock Beware of words: Evaluating the lexical richness of conversational large language models, 2024.

\bibitem{6449109}
Glenda~M. McClure.
\newblock Readability formulas: Useful or useless?
\newblock {\em IEEE Transactions on Professional Communication}, PC-30(1):12--15, 1987.

\bibitem{mcinnes2020umap}
Leland McInnes, John Healy, and James Melville.
\newblock Umap: Uniform manifold approximation and projection for dimension reduction, 2020.

\bibitem{openai2023gpt}
R~OpenAI.
\newblock Gpt-4 technical report. arxiv 2303.08774.
\newblock {\em View in Article}, 2(5), 2023.

\bibitem{opensearch}
Opensearch.
\newblock opensearch.org.

\bibitem{ouyang2022training}
Long Ouyang, Jeffrey Wu, Xu~Jiang, Diogo Almeida, Carroll Wainwright, Pamela Mishkin, Chong Zhang, Sandhini Agarwal, Katarina Slama, Alex Ray, et~al.
\newblock Training language models to follow instructions with human feedback.
\newblock {\em Advances in neural information processing systems}, 35:27730--27744, 2022.

\bibitem{pal2022medmcqa}
Ankit Pal, Logesh~Kumar Umapathi, and Malaikannan Sankarasubbu.
\newblock Medmcqa : A large-scale multi-subject multi-choice dataset for medical domain question answering, 2022.

\bibitem{pal2024domain}
Soumen Pal, Manojit Bhattacharya, Sang-Soo Lee, and Chiranjib Chakraborty.
\newblock A domain-specific next-generation large language model (llm) or chatgpt is required for biomedical engineering and research.
\newblock {\em Annals of Biomedical Engineering}, 52(3):451--454, 2024.

\bibitem{Pierrehumbert2018OnHL}
Janet~B. Pierrehumbert and Ram{\'o}n Granell.
\newblock On hapax legomena and morphological productivity.
\newblock 2018.

\bibitem{rae2021scaling}
Jack~W Rae, Sebastian Borgeaud, Trevor Cai, Katie Millican, Jordan Hoffmann, Francis Song, John Aslanides, Sarah Henderson, Roman Ring, Susannah Young, et~al.
\newblock Scaling language models: Methods, analysis \& insights from training gopher.
\newblock {\em arXiv preprint arXiv:2112.11446}, 2021.

\bibitem{rae2022scaling}
Jack~W. Rae, Sebastian Borgeaud, Trevor Cai, Katie Millican, Jordan Hoffmann, Francis Song, John Aslanides, Sarah Henderson, Roman Ring, Susannah Young, Eliza Rutherford, Tom Hennigan, Jacob Menick, Albin Cassirer, Richard Powell, George van~den Driessche, Lisa~Anne Hendricks, Maribeth Rauh, Po-Sen Huang, Amelia Glaese, Johannes Welbl, Sumanth Dathathri, Saffron Huang, Jonathan Uesato, John Mellor, Irina Higgins, Antonia Creswell, Nat McAleese, Amy Wu, Erich Elsen, Siddhant Jayakumar, Elena Buchatskaya, David Budden, Esme Sutherland, Karen Simonyan, Michela Paganini, Laurent Sifre, Lena Martens, Xiang~Lorraine Li, Adhiguna Kuncoro, Aida Nematzadeh, Elena Gribovskaya, Domenic Donato, Angeliki Lazaridou, Arthur Mensch, Jean-Baptiste Lespiau, Maria Tsimpoukelli, Nikolai Grigorev, Doug Fritz, Thibault Sottiaux, Mantas Pajarskas, Toby Pohlen, Zhitao Gong, Daniel Toyama, Cyprien de~Masson~d'Autume, Yujia Li, Tayfun Terzi, Vladimir Mikulik, Igor Babuschkin, Aidan Clark, Diego de~Las~Casas, Aurelia Guy, Chris Jones,
  James Bradbury, Matthew Johnson, Blake Hechtman, Laura Weidinger, Iason Gabriel, William Isaac, Ed~Lockhart, Simon Osindero, Laura Rimell, Chris Dyer, Oriol Vinyals, Kareem Ayoub, Jeff Stanway, Lorrayne Bennett, Demis Hassabis, Koray Kavukcuoglu, and Geoffrey Irving.
\newblock Scaling language models: Methods, analysis \& insights from training gopher, 2022.

\bibitem{reimers2019sentence}
Nils Reimers and Iryna Gurevych.
\newblock Sentence-bert: Sentence embeddings using siamese bert-networks.
\newblock {\em arXiv preprint arXiv:1908.10084}, 2019.

\bibitem{Robins1995CatastrophicFR}
Anthony~V. Robins.
\newblock Catastrophic forgetting, rehearsal and pseudorehearsal.
\newblock {\em Connect. Sci.}, 7:123--146, 1995.

\bibitem{rolnick2019experience}
David Rolnick, Arun Ahuja, Jonathan Schwarz, Timothy~P. Lillicrap, and Greg Wayne.
\newblock Experience replay for continual learning, 2019.

\bibitem{saito2024unsupervised}
Kuniaki Saito, Kihyuk Sohn, Chen-Yu Lee, and Yoshitaka Ushiku.
\newblock Unsupervised llm adaptation for question answering.
\newblock {\em arXiv preprint arXiv:2402.12170}, 2024.

\bibitem{sanh2020distilbert}
Victor Sanh, Lysandre Debut, Julien Chaumond, and Thomas Wolf.
\newblock Distilbert, a distilled version of bert: smaller, faster, cheaper and lighter, 2020.

\bibitem{singhal2023expertlevel}
Karan Singhal, Tao Tu, Juraj Gottweis, Rory Sayres, Ellery Wulczyn, Le~Hou, Kevin Clark, Stephen Pfohl, Heather Cole-Lewis, Darlene Neal, Mike Schaekermann, Amy Wang, Mohamed Amin, Sami Lachgar, Philip Mansfield, Sushant Prakash, Bradley Green, Ewa Dominowska, Blaise~Aguera y~Arcas, Nenad Tomasev, Yun Liu, Renee Wong, Christopher Semturs, S.~Sara Mahdavi, Joelle Barral, Dale Webster, Greg~S. Corrado, Yossi Matias, Shekoofeh Azizi, Alan Karthikesalingam, and Vivek Natarajan.
\newblock Towards expert-level medical question answering with large language models, 2023.

\bibitem{sinha2020impact}
Ankur Sinha and Tanmay Khandait.
\newblock Impact of news on the commodity market: Dataset and results, 2020.

\bibitem{Steck2024IsCO}
Harald Steck, Chaitanya Ekanadham, and Nathan Kallus.
\newblock Is cosine-similarity of embeddings really about similarity?
\newblock {\em ArXiv}, abs/2403.05440, 2024.

\bibitem{thoppilan2022lamda}
Romal Thoppilan, Daniel De~Freitas, Jamie Hall, Noam Shazeer, Apoorv Kulshreshtha, Heng-Tze Cheng, Alicia Jin, Taylor Bos, Leslie Baker, Yu~Du, et~al.
\newblock Lamda: Language models for dialog applications.
\newblock {\em arXiv preprint arXiv:2201.08239}, 2022.

\bibitem{toshniwal2024openmathinstruct1}
Shubham Toshniwal, Ivan Moshkov, Sean Narenthiran, Daria Gitman, Fei Jia, and Igor Gitman.
\newblock Openmathinstruct-1: A 1.8 million math instruction tuning dataset, 2024.

\bibitem{touvron2023llama}
Hugo Touvron, Thibaut Lavril, Gautier Izacard, Xavier Martinet, Marie-Anne Lachaux, Timoth{\'e}e Lacroix, Baptiste Rozi{\`e}re, Naman Goyal, Eric Hambro, Faisal Azhar, et~al.
\newblock Llama: Open and efficient foundation language models.
\newblock {\em arXiv preprint arXiv:2302.13971}, 2023.

\bibitem{wang2024role}
Rui Wang, Fei Mi, Yi~Chen, Boyang Xue, Hongru Wang, Qi~Zhu, Kam-Fai Wong, and Ruifeng Xu.
\newblock Role prompting guided domain adaptation with general capability preserve for large language models.
\newblock {\em arXiv preprint arXiv:2403.02756}, 2024.

\bibitem{wang2020minilm}
Wenhui Wang, Furu Wei, Li~Dong, Hangbo Bao, Nan Yang, and Ming Zhou.
\newblock Minilm: Deep self-attention distillation for task-agnostic compression of pre-trained transformers.
\newblock 2020.

\bibitem{wei2023chainofthought}
Jason Wei, Xuezhi Wang, Dale Schuurmans, Maarten Bosma, Brian Ichter, Fei Xia, Ed~Chi, Quoc Le, and Denny Zhou.
\newblock Chain-of-thought prompting elicits reasoning in large language models, 2023.

\bibitem{wei2023magicoder}
Yuxiang Wei, Zhe Wang, Jiawei Liu, Yifeng Ding, and Lingming Zhang.
\newblock Magicoder: Source code is all you need, 2023.

\bibitem{wu2023bloomberggpt}
Shijie Wu, Ozan Irsoy, Steven Lu, Vadim Dabravolski, Mark Dredze, Sebastian Gehrmann, Prabhanjan Kambadur, David Rosenberg, and Gideon Mann.
\newblock Bloomberggpt: A large language model for finance, 2023.

\bibitem{yang2022re3generatinglongerstories}
Kevin Yang, Yuandong Tian, Nanyun Peng, and Dan Klein.
\newblock Re3: Generating longer stories with recursive reprompting and revision, 2022.

\bibitem{10.1093/applin/amp024}
Guoxing Yu.
\newblock {Lexical Diversity in Writing and Speaking Task Performances}.
\newblock {\em Applied Linguistics}, 31(2):236--259, 06 2009.

\bibitem{zhang2023reformulating}
Yating Zhang, Yexiang Wang, Fei Cheng, Sadao Kurohashi, et~al.
\newblock Reformulating domain adaptation of large language models as adapt-retrieve-revise.
\newblock {\em arXiv preprint arXiv:2310.03328}, 2023.

\bibitem{zheng2024judging}
Lianmin Zheng, Wei-Lin Chiang, Ying Sheng, Siyuan Zhuang, Zhanghao Wu, Yonghao Zhuang, Zi~Lin, Zhuohan Li, Dacheng Li, Eric Xing, et~al.
\newblock Judging llm-as-a-judge with mt-bench and chatbot arena.
\newblock {\em Advances in Neural Information Processing Systems}, 36, 2024.

\end{thebibliography}

\newpage

\appendix
\section{Datasets} \label{datasets}
Several datasets were used in our experiments. In order to conduct ablation experiments, we used mostly Common rawl dataset. 

\begin{itemize}
\item \textbf{Common Crawl} \cite{cc}: The Common Crawl corpus contains petabytes of data collected over 12 years of web crawling. The corpus contains raw web page data, metadata extracts and text extracts. Common Crawl data is stored on Amazon Web Services’ Public Data Sets and on multiple academic cloud platforms across the world.
\end{itemize}

To train the base model we used 300B tokens and additional 100B tokens for training candidate 7B models. We tested our ablation models using several industry specific benchmark datasets. 
\begin{itemize}
\item \textbf{MMLU}\cite{hendrycks2021measuring} : MMLU stands for Multimedia Language Understanding dataset. MMLU is designed to measure knowledge acquired during pretraining by evaluating models exclusively in zero-shot and few-shot settings. The benchmark covers 57 subjects across STEM, the humanities, the social sciences, and more. It ranges in difficulty from an elementary level to an advanced professional level, and it tests both world knowledge and problem solving ability. We utilized healthcare specific datasets with in MMLU that includes anatomy, clinical knowledge, college medicine, human sexuality, medical genetics, professional medicine, and virology.
\item \textbf{MedQA - USMLE}\cite{jin2020disease}: Multiple choice question answering based on the United States Medical License Exams (USMLE). The dataset is collected from the professional medical board exams. We used the test split consisting of 1.2k QA pairs.
\item \textbf{MedMCQA}\cite{pal2022medmcqa}: MedMCQA is a large-scale, Multiple-Choice Question Answering (MCQA) dataset designed to address real-world medical entrance exam questions. We used the test split of 6.15k question answers.
\item \textbf{PubMedQA}\cite{jin2019pubmedqa}: PubMedQA is a biomedical question answering (QA) dataset collected from PubMed abstracts. The task of PubMedQA is to answer research questions with yes/no/maybe (e.g.: Do preoperative statins reduce atrial fibrillation after coronary artery bypass grafting?) using the corresponding abstracts. We used the 1k labeled test split from PubMedQA.
\item \textbf{FiQA-SA}\cite{inproceedings}: This dataset is based on the task 1 of the Financial Sentiment Analysis in the Wild (FiQA) challenge.  The dataset is split into three subsets: train, valid, test with sizes 822, 117, 234 respectively. We used the test split in our experiments.
\item \textbf{FPB}\cite{malo2013good}: The Financial PhraseBank (FPB) dataset consists of 4840 sentences from English language financial news categorised by sentiment. We used the test split consisting of 970 rows.
% \item \textbf{Headlines}\cite{sinha2020impact}: This dataset contains the gold commodity news annotated into various dimensions including information such as past movements and expected directionality in prices, asset comparison and other general information. We used the test spli consisiting of 20k samples.
\end{itemize}

\section{LLM-as-judge: Prompt} \label{sec:judge-prompt}
To assess the performance of the classifier, we also employed the LLM-as-a-judge evaluation approach outlined in \cite{zheng2024judging, huang2024empirical}. This approach leverages a large language model (LLM) to judge the industry domain predictions made by our model. We used Llama 3 \cite{dubey2024llama3herdmodels}, a state-of-the-art LLM, with a list of predicted industry domains for each document. Llama 3 \cite{dubey2024llama3herdmodels} was then asked to pass judgment on whether it agreed with the predicted domains or not. To mitigate potential position bias, where a LLM may exhibit a propensity to favor certain positions over others, we randomly shuffled the industry domain list in every prompt. The prompt used for the evaluation is shown in figure \ref{fig:judgeprompt}.

\begin{figure}[htbp]
    \centering
    \begin{subfigure}[b]{0.48\textwidth}
        \centering
        \includegraphics[width=\textwidth]{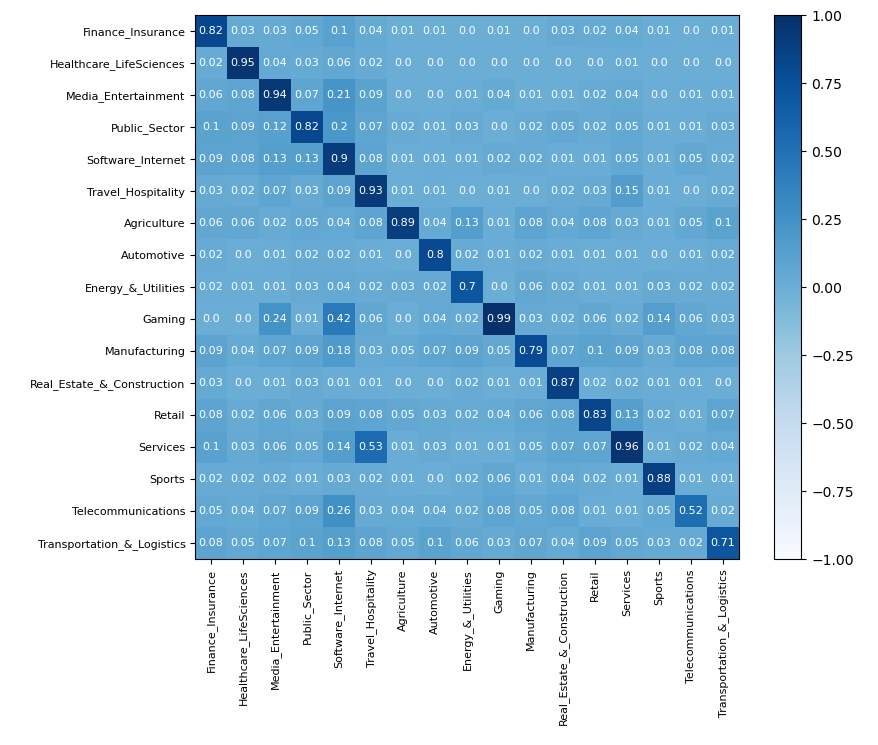}
        \caption{Classifier confusion matrix}
        \label{fig:confusion}
    \end{subfigure}
    \hfill
    \begin{subfigure}[b]{0.48\textwidth}
        \centering
        \includegraphics[width=\textwidth]{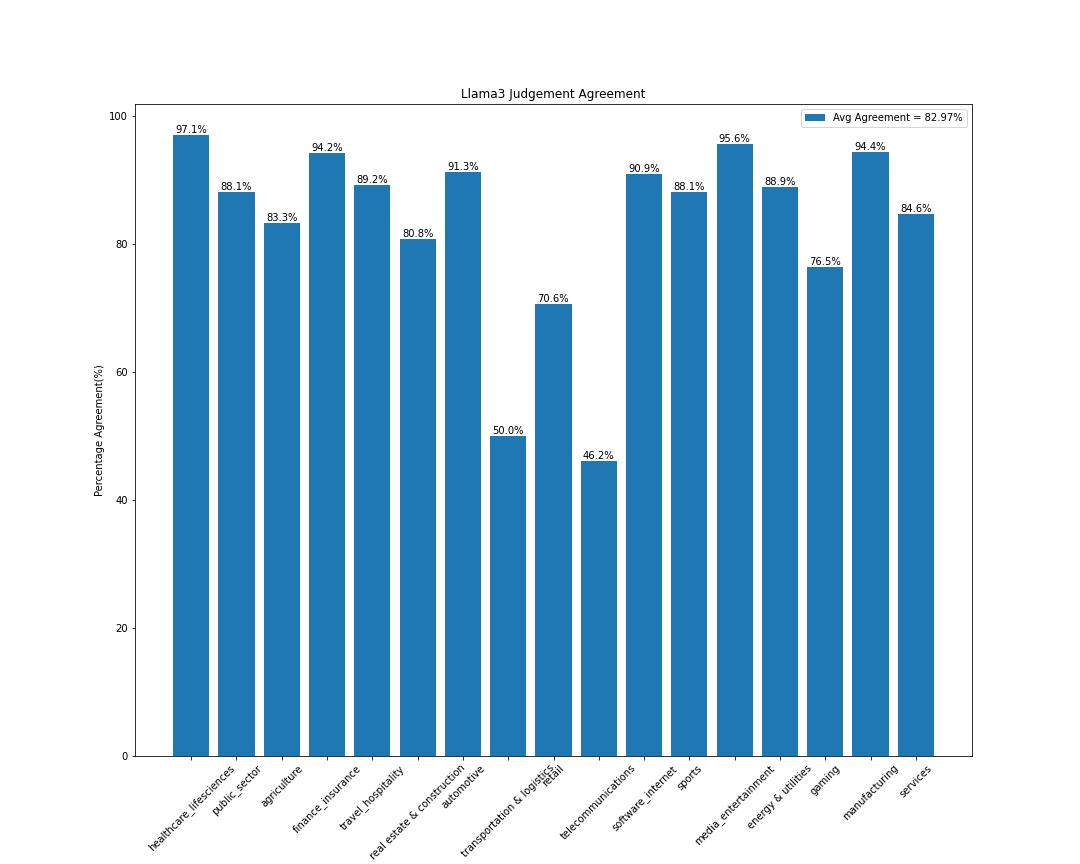}
        \caption{Agreement between LLM judge and classifier}
        \label{fig:judgeresults}
    \end{subfigure}
    \caption{Results comparison}
    \label{fig:results}
\end{figure}

\begin{figure}[!ht]
\begin{tcolorbox}[colback=green!05!white, % Background color
                  colframe=green!10!white, % Frame color
                  width=\textwidth, % Width of the tcolorbox
                  arc=2mm, % Radius of the rounded corners
                  % auto outer arc,
                  ]
% \textbf{LLM as a judge: Prompt}
\begin{verbatim}

Please act as an impartial judge and evaluate the industry domains assigned 
by a ML model to the document displayed below. The document belongs to one 
or more of the industry domains listed below:

Finance_Insurance
Software_Internet
...<truncated list>...
Telecommunications

[Document]
[Start of document]
<SOME TEXT>
[The End of document]
ML model predicted industry domains 
and respective scores:
Travel_Hospitality : 0.94
....
 
Please judge the domains assigned by the ML model  by stating if you agree
or disagree with them. Provide your reasoning for the judgement within
<COMMENTS> reasoning <\COMMENTS> tags.
Please strictly follow the below 
format for your judgement response.
<Judgement>
{'ML predicted domain' : 'rating'} 
<\Judgement>

for example:
<COMMENTS> reasoning <\COMMENTS>
<Judgement>
{'ML predicted domain': 'disagree'} 
<\Judgement>

\end{verbatim}
\end{tcolorbox}
\caption{LLM as a judge: Prompt}
\label{fig:judgeprompt}
\end{figure}

The results of this evaluation are presented in figure \ref{fig:judgeresults}, where we plot the percentage agreement between Llama 3 \cite{dubey2024llama3herdmodels} and our industry domain model predictions for each domain. The x-axis represents the industry domains, and the y-axis displays the percentage agreement with Llama 3 \cite{dubey2024llama3herdmodels}. Our analysis reveals that Llama 3 \cite{dubey2024llama3herdmodels} exhibited an overall agreement of 82.97\% on the 806 documents evaluated. However, the agreement was below 50\% for the "Transportation \& Logistics" and "Telecommunications" domains. For the remaining domains, Llama 3 \cite{dubey2024llama3herdmodels} demonstrated a high level of agreement, exceeding 80\%, with the predictions made by our industry domain classifier. This indicates that our classifier performed well in accurately assigning industry domains to the documents in our large corpus $D$.

While the LLM-as-a-judge evaluation approach provides valuable insights into the performance of our models, it is important to note that the judgments made by Llama 3 \cite{dubey2024llama3herdmodels} are subjective and may be influenced by its training data and biases. 

\section{Industry Verticals} \label{verticals}

\begin{itemize}
    \item \textbf{Financial Services \& Insurance (FSI)}:
Financial Services comprises establishments primarily engaged in financial transactions (transactions involving the creation, liquidation, or change in ownership of financial assets) and/or in facilitating financial transactions. This includes Banking, Securities and Insurance as well as the organizations that regulate or serve these institutions.
\item \textbf{Healthcare \& Life Sciences (HCLS)}:
Healthcare provides goods and services to treat patients with curative, preventive, rehabilitative, and palliative care. This includes organizations that have oversight or serve these establishments. Life Sciences encompass organizations in the fields of biotechnology, pharmaceuticals, biomedical technologies, life systems technologies, nutraceuticals, cosmeceuticals, food processing, and organizations and institutions that devote the majority of their efforts in the various stages of research, development, technology transfer and commercialization.
\item \textbf{Media \& Entertainment (M\&E)}:
The media and entertainment industry consists of film, print, radio, television, sports and cultural institutions such as libraries and museums.
\item \textbf{Public Sector}:
The Public Sector includes government agencies providing public services, prioritizing digital transformation, efficient governance, and citizen-centric solutions. It incorporates technologies like AI and smart city initiatives to enhance transparency and public welfare.
\item \textbf{Software \& Internet}:
Software \& Internet comprises organizations involved in the development, maintenance and publication of general software. The industry also includes networking and storage.
\item \textbf{Travel \& Hospitality}:
The Travel \& Hospitality sector encompasses transportation, accommodation, and leisure services, adapting to technology with online booking systems and personalized experiences. Sustainability, safety, and customer-centric approaches drive the industry to meet evolving travel preferences. 
The Hospitality industry includes lodging, event planning, theme parks, cruise line, restaurants and other establishments or services within the tourism industry.
\item \textbf{Agriculture}:
Agriculture is the science, art, or occupation concerned with cultivating land, raising crops, and feeding, breeding, and raising livestock. It also includes the production of livestock, poultry, fish, and crops. All food consumed by people and feed consumed by humans is a result of agriculture. Agricultural crops are also used for many forms of fuel.
\item \textbf{Energy \& Utilities}:
Energy: Oil and Gas: The Oil \& Gas industry comprises organizations involved in the exploration, extraction, refining, transporting, and marketing of petroleum products. This includes organizations that serve or regulate those establishments.
Energy: Power and Utilities: The Power \& Utilities industry comprises organizations involved in the generation, transmission, distribution and sale of electric, natural gas, water and other regulated utility operations. This includes organizations that serve or regulate those establishments.
\item \textbf{Gaming}:
Organizations the develop, sell or license electronic games, lotteries or contests. This includes organizations that have oversight or serve these establishments. Distribution/Platform Services, Game Developer, Game Publishing/Operations, Game Services \& Technology, Real Money Gaming, , Simulation.
\item \textbf{Real Estate \& Construction}:
Engineering, Construction \& Real Estate comprises organizations involved with the construction, alteration, engineering or sale of the physical environment. This includes organizations that serve or regulate those establishments.
\item \textbf{Retail}:
The Retail Trade sector comprises establishments engaged in retailing merchandise, generally without transformation, and rendering services incidental to the sale of merchandise. The retailing process is the final step in the distribution of merchandise; retailers are, therefore, organized to sell merchandise in small quantities to the general public. This sector comprises two main types of retailers: store and nonstore retailers.
\item \textbf{Services}:
The services industry covers a wide range of businesses providing intangible products such as consulting, finance, and healthcare. It emphasizes customer satisfaction, digital transformation, and innovative solutions, adapting to changing market demands and technological advancements.
\item \textbf{Sports}:
The sports industry involves entertainment, competition, and physical activities. It includes professional leagues, sports events, and fitness services. The industry leverages technology for fan engagement, athlete performance analysis, and sports broadcasting innovations.
\item \textbf{Automotive}:
The Automotive industry comprises organizations involved in the design, development, manufacturing, marketing, regulation and selling of motor vehicles.
\item \textbf{Manufacturing}:
The Manufacturing Industry comprises organizations involved in the fabrication, processing, or preparation of products from raw materials and commodities. This excludes items manufactured for specific industries such as communications equipment. Those organizations are classed with the industry they serve.
\item \textbf{Telecommunications}:
The Telecommunications industry comprises organizations that provide telecommunications, telephone and internet and the services or oversight related to those activities
\item \textbf{Transportation \& Logistics}:
The Transportation \& Logistics Industries are organizations involved in the transportation and storage of cargo, materials or goods by rail, air, water or road, including postal. This includes organizations that serve or regulate those establishments.
\end{itemize}

\begin{table*}[!ht]
% \centering
\caption{Seed Generation Example: Finance and Insurance domain}
\begin{tabular}{p{\textwidth}}
\hline
\textbf{Seed Generation Example: Finance and Insurance domain}  \\
\hline
\textbf{Prompt:}
\begin{Verbatim}[commandchars=+\[\]]
Write a +underline[legal brief] about +underline[Financial Services and Insurance] using the following steps.
1. Generate a random topic from +underline[Financial Services and Insurance] domain.
2. Write a short premise for the +underline[legal brief] about the topic from 
+underline[Financial Services and Insurance] 
3. Write a short description of the author of the +underline[legal brief]. The author should be a 
practicing member of the +underline[Financial Services and Insurance] industry.
4. Describe the +underline[legal brief]'s audience.
5. Give the author's motive in writing the document for the audience. The author's 
demeanor is +underline[professional].
6. Write a +underline[legal brief] about +underline[Financial Services and Insurance] based on the 
topic generated, premise from the perspective and motive of the author 
targeting to the audience.
The resulting document should be +underline[very long (more than 1500 words)]. Your response 
should be in the following format.
        - TOPIC: 
        - PREMISE: 
        - AUTHOR:
        - AUDIENCE:
        - MOTIVE: 
        - DOCUMENT:
\end{Verbatim}
% \hline
\textbf{Generation:}
\begin{Verbatim}[commandchars=+\[\]]
- TOPIC:Cryptocurrency Regulations in the Insurance Industry
- PREMISE: The legal brief aims to provide an overview of the current regulatory 
landscape surrounding the use of cryptocurrencies in the insurance industry, 
highlighting the potential risks and challenges, as well as the opportunities that 
this emerging technology presents.
- AUTHOR: Sarah Johnson is a seasoned legal professional with over 15 years of 
experience in the financial services and insurance sector. She currently serves as 
the General Counsel for a leading insurance company, where she oversees all legal 
matters related to compliance, risk management, and regulatory affairs.
- AUDIENCE: The legal brief is intended for senior executives, board members, 
and key stakeholders within the insurance industry, as well as regulatory bodies 
and policymakers involved in shaping the legal framework surrounding cryptocurrencies.
- MOTIVE: Sarah Johnson's primary motive in writing this legal brief is to provide a 
comprehensive and impartial analysis of the current regulatory environment surrounding 
cryptocurrencies in the insurance industry. As a seasoned legal professional, she 
recognizes the importance of staying ahead of emerging trends and technologies that 
could potentially disrupt the industry. By offering an in-depth examination of the legal
and regulatory challenges associated with cryptocurrencies, Sarah aims to equip 
decision-makers with the necessary knowledge to navigate this complex landscape and make
informed decisions that protect their organizations while embracing innovative solutions.
- DOCUMENT: Introduction:
The rapid rise of cryptocurrencies has disrupted numerous industries, and the insurance
sector is no exception. As these digital assets gain traction and mainstream adoption, 
insurers mustgrapple with the legal and regulatory implications of incorporating 
cryptocurrencies into their operations. This legal brief seeks to provide...<truncated>
\end{Verbatim}
% \hline
\\
\bottomrule
\end{tabular}

\end{table*}
\begin{table*}[htbp]
\centering
\caption{Seed Generation Example: Healthcare and Life Sciences domain}
\begin{tabular}{p{\textwidth}}
\hline
\textbf{Seed Generation Example: Healthcare and Life Sciences domain}  \\
\hline
\textbf{Prompt:}
\begin{Verbatim}[commandchars=+\[\]]
Write a +underline[product proposal] about +underline[Healthcare & Life Sciences] using
the following steps.
1. Generate a random topic from Healthcare & Life Sciences domain.
2. Write a short premise for the +underline[product proposal] about the topic from 
+underline[Healthcare & Life Sciences] 
3. Write a short description of the author of the +underline[product proposal]. The 
author should be a practicing member of the +underline[Healthcare & Life Sciences] 
industry.
4. Describe the +underline[product proposal]'s audience.
5. Give the author's motive in writing the document for the audience. The author's 
demeanor is +underline[professional].
6. Write a +underline[product proposal] about +underline[Healthcare & Life Sciences]
based on the topic generated, premise from the perspective and motive of the author 
targeting to the audience.
The resulting document should be +underline[short (less than 500 words)].
Your response should be in the following format.
        - TOPIC: 
        - PREMISE: 
        - AUTHOR:
        - AUDIENCE:
        - MOTIVE: 
        - DOCUMENT:
\end{Verbatim}
% \hline
\textbf{Generation:}
\begin{Verbatim}[commandchars=+\[\]]
- TOPIC: Wearable Health Monitoring Devices for Elderly Care
- PREMISE: The product proposal aims to introduce a new line of wearable health 
monitoring devices specifically designed for elderly individuals, enabling real-time
tracking of vital signs and promoting proactive healthcare management.
- AUTHOR: Dr. Emily Thompson is a renowned geriatrician with over 20 years of 
experience in the healthcare industry. She currently serves as the Chief Medical 
Officer at a leadingv assisted living facility, where she oversees the medical 
care and well-being of elderly residents.
- AUDIENCE: The product proposal is targeted towards healthcare providers, 
caregivers, and decision-makers in the elderly care sector, including nursing 
home administrators, assisted living facility managers, and healthcare technology
companies.
- MOTIVE: Dr. Emily Thompson's primary motive in writing this product proposal 
is to address the growing demand for innovative healthcare solutions tailored to
the unique needs of the aging population. As a seasoned geriatrician, she recognizes
the importance of proactive health monitoring and early intervention in reducing
the risk of hospitalizations and improving the overall quality of life for elderly
individuals. By introducing a new line of wearable health monitoring devices,
Dr. Thompson aims to empower caregivers and healthcare providers with real-time 
data and actionable insights, ultimately enhancing the standard ofcare for elderly
patients.
- DOCUMENT: Dear Healthcare Professionals and Decision-Makers,

As our population continues to age, the demand for innovative healthcare solutions
tailored to the unique needs of elderly individuals has never been greater. 
In response to this pressing need, I am excited to introduce...<truncated>

\end{Verbatim}
% \hline
\\
\bottomrule
\end{tabular}
\end{table*}

\begin{table*}[!http]
\centering
\caption{Seed Generation Example for multiple domains: Sports and Travel/Hospitality}
\label{tab:multiple}
\begin{tabular}{p{\textwidth}}
\hline
\textbf{Seed Generation Example for multiple domains: Sports and Travel/Hospitality}  \\
\hline
\textbf{Prompt:}
\begin{Verbatim}[commandchars=+\[\]]
Write a +underline[textbook chapter] about 2 industries: +underline[Sports and Travel/Hospitality]
using the following steps.
1. Generate a random topic from +underline[Sports and Travel/Hospitality] domains
2. Write a short premise(less than 30 words) for the +underline[textbook chapter] about the topic
from +underline[Sports and Travel/Hospitality] domains.
3. Write a short description(less than 30 words) of the author of the +underline[textbook chapter].
4. Describe the +underline[textbook chapter]’s audience.
5. Give the author's motive(less than 30 words) in writing the document for the audience. 
The author's demeanor is professional.
6. Write a +underline[textbook chapter] about +underline[Sports and Travel/Hospitality] industries
based on the topic generated, premise from the perspective and motive of the author.
The resulting document should be very long (more than 1500 words). Your response should
be in the following format.
    - TOPIC: 
    - PREMISE: 
    - AUTHOR:
    - AUDIENCE
    - MOTIVE: 
    - DOCUMENT: 
\end{Verbatim}
% \hline
\textbf{Generation:}
\begin{Verbatim}[commandchars=+\[\]]
- TOPIC: The Fusion of Sports and Tourism: Exploring the Rise of Sports Tourism
- PREMISE: Examine the growing phenomenon of sports tourism and its impact on the 
travel industry.
- AUTHOR: Dr. Michael Thompson, a seasoned tourism researcher with over 15 years of 
experience in the field.
- AUDIENCE: Industry professionals and policymakers.
- MOTIVE: To provide a comprehensive analysis of sports tourism for industry 
professionals and policymakers.
- DOCUMENT:

Chapter 1: The Fusion of Sports and Tourism: Exploring the Rise of Sports Tourism

In recent years, the worlds of sports and tourism have converged, giving rise to a 
unique and rapidly growing phenomenon known as sports tourism. This fusion has opened 
up new opportunities for destinations, athletes, and fans alike, creating a dynamic 
industry that combines the thrill of athletic competition with the allure of travel 
and cultural exploration.

Section 1: Understanding Sports Tourism

Sports tourism can be defined as the practice of traveling to participate in or 
spectate at sporting events, competitions, or activities. It encompasses a wide 
range of activities, from attending major international sporting events like the Olympics
or the FIFA World Cup, to participating in recreational activities such as golf, skiing, 
or adventure sports...

<truncated>

\end{Verbatim}
% \hline
\\
\bottomrule
\end{tabular}
\end{table*}
\end{document}